\documentclass[final]{cvpr}

\usepackage{times}
\usepackage{epsfig}
\usepackage{graphicx}
\usepackage{amsmath}
\usepackage{amssymb}

\usepackage{siunitx}
\usepackage{cite}
\usepackage{ctable}
\usepackage{hhline}
\usepackage{booktabs}
\usepackage{subcaption}
\usepackage{csquotes}
\usepackage{pifont}
\usepackage{mathtools}
\usepackage{animate}
\newlength{\itemwidth}
\usepackage{anyfontsize}
\usepackage[super]{nth}
\usepackage{multirow}

\usepackage{soul}
\usepackage{multicol}
\usepackage{makecell}
\usepackage[normalem]{ulem}
\usepackage{multirow}
\usepackage{booktabs}
\usepackage{color, colortbl}
\definecolor{Gray}{gray}{0.9}
\usepackage[pagebackref=true,breaklinks=true,letterpaper=true,colorlinks,bookmarks=false]{hyperref}

\newcommand{\floor}[1]{\left\lfloor #1 \right\rfloor}

\newcolumntype{L}[1]{>{\raggedright\let\newline\\\arraybackslash\hspace{0pt}}m{#1}}
\newcolumntype{C}[1]{>{\centering\let\newline\\\arraybackslash\hspace{0pt}}m{#1}}
\newcolumntype{R}[1]{>{\raggedleft\let\newline\\\arraybackslash\hspace{0pt}}m{#1}}

\newcommand{\cmark}{\ding{51}}%
\newcommand{\xmark}{\ding{55}}%

\usepackage[pagebackref=true,breaklinks=true,colorlinks,bookmarks=false]{hyperref}

\def\cvprPaperID{1013} % *** Enter the CVPR Paper ID here
\def\confYear{CVPR 2021}
\title{Beyond Static Features for \\ Temporally Consistent 3D Human Pose and Shape from a Video}
\author{
Hongsuk Choi$^1$\hspace{1.0cm} Gyeongsik Moon$^1$\hspace{1.0cm} Ju Yong Chang$^2$\hspace{1.0cm} Kyoung Mu Lee$^1$\\
\\
$^{1}$ECE \& ASRI, Seoul National University, Korea \hspace{1.0cm}
$^{2}$ECE, Kwangwoon University, Korea\\
{\small \texttt {\{redarknight,mks0601,kyoungmu\}@snu.ac.kr},  \texttt {juyong.chang@gmail.com}}
}

\begin{document}

\twocolumn[{
\maketitle
% \vspace{-2.5em}
\begin{center}\centering
\setlength{\tabcolsep}{0.1cm}
\setlength{\itemwidth}{5.0cm}
\hspace*{-\tabcolsep}
\begin{tabular}{ccc}

\animategraphics[width=0.896\itemwidth, poster=29, loop, autoplay, final, nomouse, method=widget]{20}{fig/teaser/input/}{000045}{000107}
&
\animategraphics[width=\itemwidth, poster=29, loop, autoplay, final, nomouse, method=widget]{20}{fig/teaser/ours/}{000045}{000107}
&
\animategraphics[width=\itemwidth, poster=29, loop, autoplay, final, nomouse, method=widget]{20}{fig/teaser/vibe/}{000045}{000107}
% \includegraphics[width=\itemwidth]{main/fig/teaser/vibe/000074.jpg}

% 74가 poster로 좋은듯

\vspace{0.1cm} \\

input

&
\textbf{TCMR (Ours)} 

&
VIBE

\end{tabular}

%\vspace{-1em}
\captionof{figure}
{
VIBE~\cite{kocabas2020vibe}, the state-of-the-art video-based 3D human pose and shape estimation method, outputs very different 3D human poses per frame, although the frames have subtle differences.
Our TCMR produces clearly more temporally consistent and smooth 3D human motion.
% \emph{To watch the video, please refer to our arxiv paper.}
\emph{This is a video figure that is best viewed by Adobe Reader.}
}
\label{fig:intro_comparision_vibe}
\end{center}
\vspace{1em}
}]

\maketitle

%%%%%%%%% ABSTRACT
\begin{abstract}
% \vspace{-5mm}
Despite the recent success of single image-based 3D human pose and shape estimation methods, recovering temporally consistent and smooth 3D human motion from a video is still challenging. 
Several video-based methods have been proposed; however, they fail to resolve the single image-based methods’ temporal inconsistency issue due to a strong dependency on a static feature of the current frame. 
In this regard, we present a temporally consistent mesh recovery system (TCMR). 
It effectively focuses on the past and future frames’ temporal information without being dominated by the current static feature.
Our TCMR significantly outperforms previous video-based methods in temporal consistency with better per-frame 3D pose and shape accuracy.
We also release the codes\footnote{\url{https://github.com/hongsukchoi/TCMR_RELEASE}}.
\end{abstract}

\vspace{-1em}

\section{Introduction}~\label{sec:introduction}
Various methods have been proposed to analyze humans from images, ranging from estimating a simplistic 2D skeleton to recovering 3D human pose and shape. 
Despite the recent improvements, estimating 3D human pose and shape from images is still a challenging task, especially in the monocular case due to depth ambiguity, limited training data, and complexity of human articulations.

Most of the previous methods~\cite{kanazawa2018end,pavlakos2018learning,kolotouros2019convolutional,kolotouros2019learning,moon2020i2l,choi2020p2m} attempt to recover 3D human pose and shape from a single image. 
They are generally based on parametric 3D human mesh models, such as SMPL~\cite{loper2015smpl}, and directly regress the model parameters from the input image.
Although single image-based methods predict a reasonable output from a static image, they tend to produce temporally inconsistent and unsmooth 3D motion when applied to a video per frame. 
The temporal instability is from inconsistent 3D pose errors for consecutive frames.
For example, the errors could occur in different 3D directions, or the following frames' pose outputs could remain relatively the same, not reflecting the motion.

Several methods~\cite{kanazawa2019learning,kocabas2020vibe,luo20203d} have been proposed to extend the single image-based methods to the video case effectively.
They feed a sequence of images to the pretrained single image-based 3D human pose and shape estimation networks~\cite{kanazawa2018end,kolotouros2019learning} to obtain a sequence of static features.
All input frames' static features are passed to a temporal encoder, which encodes a temporal feature for each input frame.
Then, a body parameter regressor outputs SMPL parameters for each frame from the temporal feature of the corresponding time step.

Although the above works quantitatively improved the per-frame 3D pose accuracy and motion smoothness, their qualitative results still suffer from the temporal inconsistency aforementioned, as shown in Figure~\ref{fig:intro_comparision_vibe}.
We argue that the failure comes from a strong dependency on the static feature of the \emph{current} frame.
For terminological convenience, we use a word \emph{current} to indicate the time step of a target frame where SMPL parameters to be estimated.
The first reason for the strong dependency is a residual connection between the current frame's static and temporal features.
While the residual connection has been widely verified to facilitate a learning process, naively applying it to the temporal encoding can hinder the system from learning useful temporal information.
Given that the static feature is extracted by the pretrained network~\cite{kanazawa2018end,kolotouros2019learning}, it contains a strong cue for the SMPL parameters of the current frame.
Thus, the residual connection's identity mapping of the static feature can make the SMPL parameter regressor heavily depend on it and leverage the temporal feature marginally.
This procedure can constrain the temporal encoder from encoding more meaningful temporal features.
The second reason is the temporal encoding that takes static features from all frames, which include a current static feature.
The current static feature has the largest potential to affect the current temporal feature, from which SMPL parameters are predicted.
This phenomenon is caused by the current static feature having the most crucial information for 3D human pose and shape of a current frame.
Although the dominance will increase the per-frame accuracy of 3D pose and shape estimation, it can prevent the temporal encoder from fully exploiting the past and future frames’ temporal information. 
Taken together, the existing video-based methods have a strong preference for the current static feature, and suffer from the temporal inconsistency issue as single image-based methods do.

In this work, we propose a temporally consistent mesh recovery system (TCMR).
It is designed to resolve the strong dependency on the current static feature for temporally consistent and smooth 3D human motion output from a video.
First, although we follow the previous video-based works~\cite{kocabas2020vibe,luo20203d,kanazawa2019learning} to encode a temporal feature of the current frame, we remove the residual connection between the static and temporal features.
Moreover, we introduce PoseForecast, which consists of two temporal encoders, to \emph{forecast} a current pose from the past and future frames without the current frame.
The temporal features from PoseForecast are free from the current static feature; however, they contain essential temporal information of the past and future frames to \emph{forecast} a current pose.
The temporal features from PoseForecast are integrated with the current temporal feature, which is extracted from all input frames, to predict current SMPL parameters.
The parameters estimated from the integrated temporal feature are the final output in inference time.
By removing the strong dependency on the current static feature, our SMPL parameter regressor can have more chance to focus on the past and future frames without being dominated by the current frame.

Despite its simplicity, we observed that our newly designed temporal architecture is highly effective on obtaining the temporally consistent and smooth 3D human motion.
It also improves the accuracy of the 3D pose and shape per frame by utilizing better temporal information. 
We show that the proposed TCMR outperforms the previous video-based methods~\cite{kocabas2020vibe,luo20203d,kanazawa2019learning} on various 3D video benchmarks, especially in temporal consistency.

Our contributions can be summarized as follows.
\begin{itemize}
\item We present a temporally consistent mesh recovery system (TCMR), which produces temporally consistent and smooth 3D human motion from a video.
It effectively leverages temporal information from the past and future frames without being dominated by the static feature of the current frame.
\item Despite its simplicity, TCMR not only improves the temporal consistency of 3D human motion but also increases per-frame 3D pose and shape accuracy compared to a baseline method.
\item TCMR outperforms previous video-based methods in temporal consistency by a large margin while achieving better per-frame 3D pose and shape accuracy.
\end{itemize}

\section{Related works} 

\noindent\textbf{Single image-based 3D human pose and shape estimation.}
Most of the current single image-based 3D human pose and shape estimation methods are based on the model-based approach, which predicts parameters of a predefined 3D human mesh model, SMPL~\cite{loper2015smpl}.
Kanazawa~\etal~\cite{kanazawa2018end} proposed an end-to-end trainable human mesh recovery (HMR) system that uses adversarial loss to make their output 3D human mesh anatomically plausible. 
Pavlakos~\etal~\cite{pavlakos2018learning} used 2D joint heatmaps and silhouette as cues for predicting accurate SMPL parameters. 
Omran~\etal~\cite{omran2018neural} proposed a similar system, which uses human part segmentation as a cue for regressing SMPL parameters. 
Pavlakos~\etal~\cite{pavlakos2019texturepose} proposed a system that uses multi-view color consistency to supervise a network using multi-view geometry.
Kolotouros~\etal~\cite{kolotouros2019learning} introduced a self-improving system that consists of an SMPL parameter regressor and an iterative fitting framework~\cite{bogo2016keep}. 
Georgakis~\etal~\cite{georgakis2020hierarchical} incorporated hierarchical kinematic prior on a human body to a network.

\begin{figure*}
\begin{center}
\includegraphics[width=1.0\linewidth]{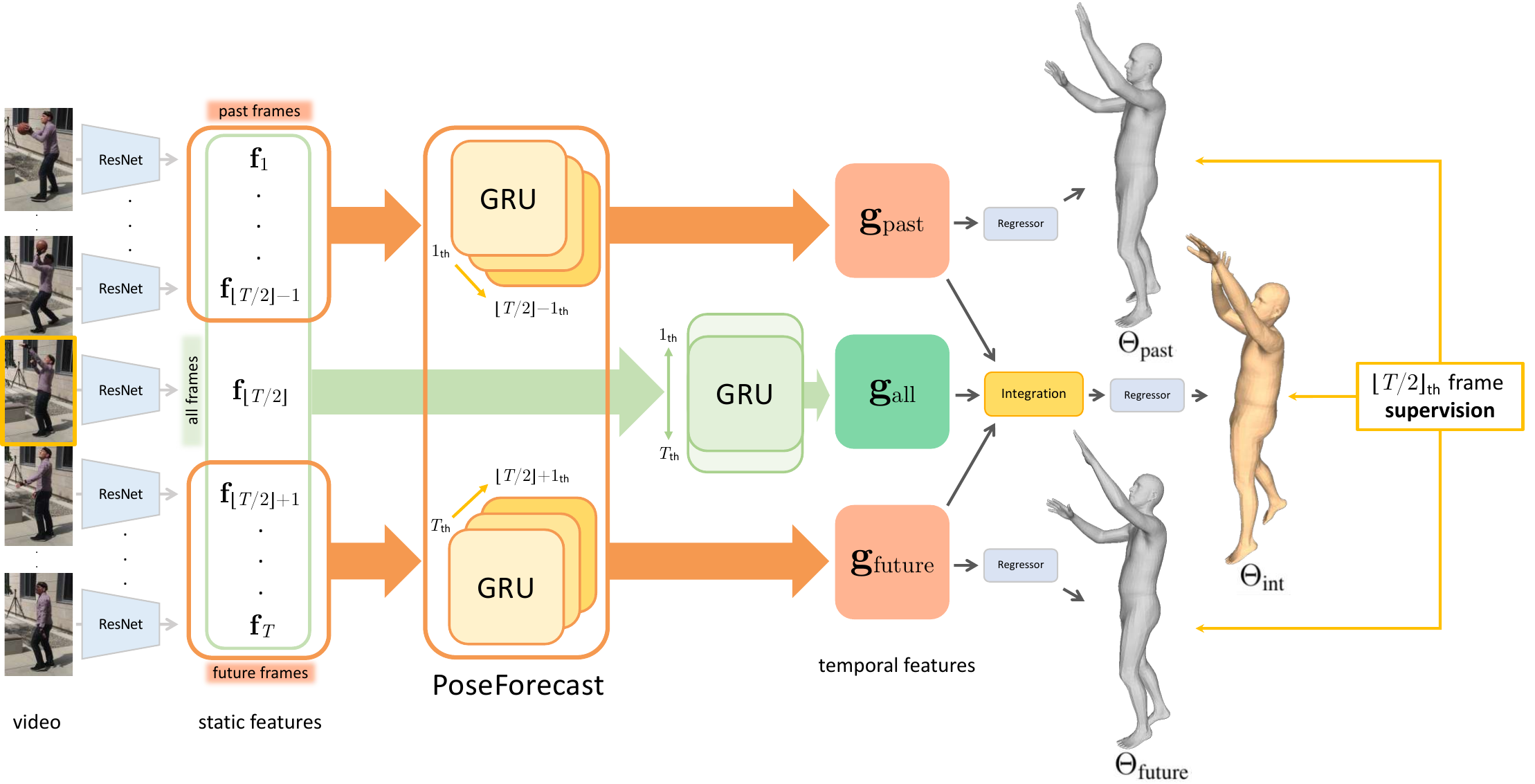}
\end{center}
\vspace*{-5mm}
   \caption{
   The overall pipeline of TCMR. The gold-colored output $\Theta_\text{int}$ is used in inference time, which is regressed from the integrated temporal feature.
   }
\vspace*{-3mm}
\label{fig:overall_pipeline}
\end{figure*}

Conversely, the model-free approach estimates the shape directly instead of regressing the model parameters. 
Varol~\etal~\cite{varol2018bodynet} proposed BodyNet, which estimates 3D human shape in the 3D volumetric space. 
Kolotouros~\etal~\cite{kolotouros2019convolutional} designed a graph convolutional human mesh regression system.
Their graph convolutional network takes a template human mesh in a rest pose as input and predicts mesh vertex coordinates using image features from ResNet~\cite{he2016deep}.
Moon and Lee~\cite{moon2020i2l} introduced a lixel-based 1D heatmap to localize mesh vertices in a fully convolutional manner.
Choi~\etal~\cite{choi2020p2m} proposed a graph convolutional network that recovers 3D human pose and mesh from a 2D human pose.

Despite moderate performance on a static image, the single image-based works suffer from temporal inconsistency (\textit{e.g.}, sudden change of poses), when applied to a video.

\noindent\textbf{Video-based 3D human pose and shape estimation.}
HMMR~\cite{kanazawa2019learning} extracts static features and encodes them to a temporal feature using a 1D fully convolutional temporal encoder.
It learns temporal context representation to reduce the 3D prediction's temporal inconsistency by predicting 3D poses in the nearby past and future frames. 
% It employs the adversarial loss for the plausible human pose.
Doersch~\etal~\cite{doersch2019sim2real} trained their network on a sequence of optical flow and 2D poses to make their network generalize well to unseen videos.
Sun~\etal~\cite{sun2019human} proposed a skeleton-disentangling framework, which separates 3D human pose and shape estimation into multi-level spatial and temporal subproblems.
They enforced the network to order shuffled frames to encourage temporal feature learning.
VIBE~\cite{kocabas2020vibe} encodes static features from the input frames into a temporal feature by using a bi-directional gated recurrent unit (GRU)~\cite{cho2014learning}, and feeds it to an SMPL parameter regressor.
A motion discriminator is introduced to encourage the regressor to produce plausible 3D human motion.
MEVA~\cite{luo20203d} addresses the problem in a coarse-to-fine manner.
Their system initially estimates the coarse 3D human motion using a variational motion estimator (VME), and predicts the residual motion with a motion residual regressor (MRR).

\noindent\textbf{Temporally consistent 3D human motion from a video.}
Although there have been many methods for video-based 3D human motion estimation~\cite{rayat2018exploiting,cai2019exploiting,pavllo20193d,mehta2017vnect,kanazawa2019learning,kocabas2020vibe,luo20203d,doersch2019sim2real,sun2019human}, most of them showed their results only qualitatively, and did not report numerical evaluation on temporal consistency.
After the HMMR~\cite{kanazawa2019learning} introduced the 3D pose acceleration error for the temporal consistency and smoothness of human motion, the following works~\cite{kocabas2020vibe,luo20203d} have reported the error metric.
HMMR and VIBE~\cite{kocabas2020vibe} lowered the acceleration error compared with the single image-based methods.
However, they revealed a trade-off between per-frame accuracy and temporal consistency.
The HMMR outputs smoother 3D human motion but provides low per-frame 3D pose accuracy.
Conversely, the VIBE~\cite{kocabas2020vibe} shows high per-frame 3D pose accuracy; however, the output is temporally inconsistent in quantitative metrics and qualitative results compared with HMMR.

In this regard, MEVA~\cite{luo20203d} attempts to establish the balance between the per-frame 3D pose accuracy and the temporal smoothness. 
Although it provides better results in both metrics, the qualitative results still expose unsmooth 3D motion.
The reason is that the system strongly depends on the current static feature to estimate the current 3D pose and shape.
First, MEVA uses a residual connection between the current frames' static and temporal features.
In addition, the current temporal feature, which is used to refine initial 3D pose and shape by MRR, is encoded from static features of all frames, which include the current frame.
This procedure can make the temporal feature dominated by the current static feature.
As a result, the refinement is significantly driven by the current static feature, and the 3D errors from consecutive frames appear inconsistent.
On the contrary, our TCMR is deliberately designed to reduce the strong dependency on the static feature.
The residual connection is removed, and PoseForecast forecasts additional temporal features from past and future frames without a current frame.
%We experimentally demonstrate that our approach of alleviating the dependency increases the temporal consistency and per-frame accuracy in various ways.
Our approach alleviates the dependency and provides temporally consistent and accurate 3D human motions in both qualitative and quantitative manners.

\noindent\textbf{Forecasting 3D human poses from images.}
Recently,~\cite{chao2017forecasting,kanazawa2019learning,yuan2019ego,zhang2019predicting} proposed to predict a person's future 3D human poses from RGB input.
Chao~\etal~\cite{chao2017forecasting} leveraged a recurrent neural network (RNN) to forecast a sequence of 2D poses from a static image, and estimate 3D poses from the predicted 2D poses.
The HMMR~\cite{kanazawa2019learning} predicts the current, future, and past 3D poses from a current input image using a hallucinator.
It hallucinates the past and future 3D poses from a current frame and is self-supervised by the output of the 1D fully convolutional temporal encoder. 
Zhang~\etal~\cite{zhang2019predicting} proposed a neural autoregressive framework that takes past video frames as input to predict future 3D motion.
Yuan~\etal~\cite{yuan2019ego} adopted deep reinforcement learning to forecast future 3D human poses from egocentric videos.
Although the objective of the above methods is to forecast future 3D poses, our system aims to learn useful temporal features free from a current static feature by the forecasting.

\vspace*{-1em}
\section{TCMR}
Figure~\ref{fig:overall_pipeline} shows the overall pipeline of our TCMR.
We provide descriptions of each part in the system as follows.

\subsection{Temporal encoding from all frames}
Given a sequence of $T$ RGB frames $\mathbf{I}_1, \dots, \mathbf{I}_T$, ResNet~\cite{he2016deep}, pretrained by Kolotouros~\etal~\cite{kolotouros2019learning}, extracts a static image feature per frame.
Then, a global average pooling is applied on the ResNet outputs, which become $\mathbf{f}_1, \dots, \mathbf{f}_T$, where $\mathbf{f}_\bullet \in \mathbb{R}^{2048}$.  
The network weights of the ResNet are shared for all frames.

From the extracted static features of all input frames, we compute the current frame's temporal feature using a bi-directional GRU, which consists of two uni-directional GRUs.
We denote the bi-directional GRU as $\mathcal{G}_\text{all}$.
The current frame is defined as a $\floor{T/2}$th frame among $T$ input frames.
The two uni-directional GRUs extract temporal features from the input static features in the opposite time directions.
% the past and future frames, where past and future frames are defined as $1, \dots, \floor{T/2}-1$th frames and $\floor{T/2}+1, \dots, T$th frames, respectively.
The initial inputs of the two GRUs are $\mathbf{f}_1$ and $\mathbf{f}_T$, respectively, and the initial hidden states of them are initialized as zero tensors.
Then, they recurrently updates their hidden states by aggregating the static features from the next frames $\mathbf{f}_2, \dots, \mathbf{f}_{\floor{T/2}}$ and $\mathbf{f}_{T-1}, \dots, \mathbf{f}_{\floor{T/2}}$, respectively.
The concatenated hidden states of the GRUs at the current frame become the current temporal feature from all input frames $\mathbf{g}_\text{all} \in \mathbb{R}^{2048}$.
Unlike VIBE~\cite{kocabas2020vibe}, we do not add residual connection between $\mathbf{f}_{\floor{T/2}}$ and $\mathbf{g}_\text{all}$, such that the current temporal feature will not be dominated by $\mathbf{f}_{\floor{T/2}}$.

\begin{figure}
\begin{center}
\includegraphics[width=1.0\linewidth]{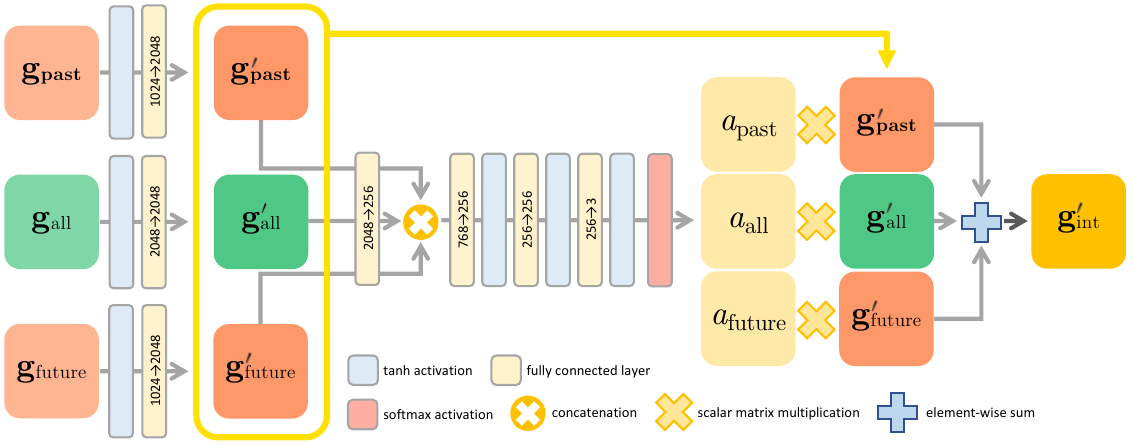}
\end{center}
\vspace*{-5mm}
\caption{
Temporal feature integration to estimate 3D human mesh for the current frame.
}
\vspace*{-3mm}
\label{fig:temporal_feature_int}
\end{figure}

% \subsection{Temporal feature only from the past and future frames}
\subsection{Temporal encoding by PoseForecast}
PoseForecast \emph{forecasts} additional temporal features for the current target pose from the past and future frames by employing two additional GRUs, denoted as $\mathcal{G}_\text{past}$ and $\mathcal{G}_\text{future}$, respectively.
The past and future frames are defined as $1, \dots, (\floor{T/2}-1)$th frames and $(\floor{T/2}+1), \dots, T$th frames, respectively.
The initial input of the $\mathcal{G}_\text{past}$ is $\mathbf{f}_1$, and the initial hidden state is initialized as a zero tensor.
Then, it recurrently updates its hidden state by aggregating the static features from the next frames $\mathbf{f}_2, \dots, \mathbf{f}_{\floor{T/2}-1}$.
The final hidden state of the $\mathcal{G}_\text{past}$ becomes the temporal feature from the past frames $\mathbf{g}_\text{past} \in \mathbb{R}^{1024}$.
Similarly, $\mathcal{G}_\text{future}$ takes $\mathbf{f}_T$ as an initial input with a zero-initialized hidden state, and recurrently updates its hidden state by aggregating the static features from the next frames $\mathbf{f}_{T-1}, \dots, \mathbf{f}_{\floor{T/2}+1}$.
The final hidden state of the $\mathcal{G}_\text{future}$ becomes the temporal feature from the future frames $\mathbf{g}_\text{future} \in \mathbb{R}^{1024}$.

\subsection{Temporal feature integration}
We integrate the extracted temporal features from all frames $\mathbf{g}_\text{all}$, from the past frames $\mathbf{g}_\text{past}$, and from the future frames $\mathbf{g}_\text{future}$ for the final 3D mesh estimation, as illustrated in Figure~\ref{fig:temporal_feature_int}.
For the integration, we pass each temporal feature to ReLU activation function and a fully connected layer to change the size of the channel dimension to 2048.
The outputs of the fully connected layer are denoted as $\mathbf{g}'_\text{all}$, $\mathbf{g}'_\text{past}$, and $\mathbf{g}'_\text{future}$.
Then, the output features are resized to 256 by a shared fully connected layer and concatenated.
The concatenated feature is passed to several fully connected layers, followed by the softmax activation function, which produces attention values $\mathbf{a}=(a_\text{all}, a_\text{past}, a_\text{future}) \in \mathbb{R}^3$.
The attention values represent how much the system should give a weight for the feature integration.
The final integrated temporal feature is obtained by $\mathbf{g}'_\text{int}=a_\text{all}\mathbf{g}'_\text{all} + a_\text{past}\mathbf{g}'_\text{past} + a_\text{future}\mathbf{g}'_\text{future}$.

In the training stage, we pass $\mathbf{g}'_\text{past}$, $\mathbf{g}'_\text{future}$, and $\mathbf{g}'_\text{int}$ to the SMPL parameter regressor, which outputs $\Theta_\text{past}$, $\Theta_\text{future}$, and $\Theta_\text{int}$ from each input temporal feature, respectively.
The regressor is shared for all outputs. 
$\Theta_\bullet$ denotes a union of SMPL parameter set $\{\theta_\bullet, \beta_\bullet\}$ and weak-perspective camera parameter set $\{s_\bullet, t_\bullet\}$.
$\theta$, $\beta$, $s$, and $t$  represent SMPL pose parameter, identity parameter, scale, and translation, respectively.
% During training, we supervise all three $\Theta_\bullet$ with current frame groundtruth.
In the testing stage, we only pass $\mathbf{g}'_\text{int}$ to the parameter regressor and use $\Theta_\text{int}$ as the final 3D human mesh.

\subsection{Loss functions}

% For the training, we calculate loss functions on $\Theta_\text{past}$, $\Theta_\text{future}$, and $\Theta_\text{int}$.
For the training, we supervise all three outputs $\Theta_\text{past}$, $\Theta_\text{future}$, and $\Theta_\text{int}$ with current frame groundtruth.
$L2$ loss between predicted and groundtruth SMPL parameters and 2D/3D joint coordinates are used, following VIBE~\cite{kocabas2020vibe}.
The 3D joint coordinates are obtained by forwarding the SMPL parameters to the SMPL layer, and the 2D joint coordinates are obtained by projecting the 3D joint coordinates using the predicted camera parameters.
% All the groundtruths are from the current frame.
\section{Implementation details}

Following VIBE~\cite{kocabas2020vibe}, we set the length of the input sequence $T$ to 16 and the input video frame rate to 25-30 frames per second and initialize the backbone and regressor with the pretrained SPIN~\cite{kolotouros2019learning}.
The weights are updated by the Adam optimizer~\cite{kingma2014adam} with a mini-batch size of 32. 
The human body region is cropped using a groundtruth box in both of training and testing stages following previous works~\cite{kanazawa2018end,kolotouros2019convolutional,kolotouros2019learning,kocabas2020vibe}.
The cropped image is resized to 224$\times$224.
Inspired by Sarandi~\etal~\cite{sarandi2018robust}, we occlude the cropped image with various objects for data augmentation.
The occlusion augmentation reduces both pose and acceleration errors approximately by 1$mm$.
Following~\cite{kocabas2020vibe,kanazawa2019learning}, we precompute the static features from the cropped images by ResNet~\cite{he2016deep} to save training time and memory.
All the 3D rotations of $\theta$ are initially predicted in the 6D rotational representation of Zhou~\etal~\cite{zhou2019continuity}, and converted to the 3D axis-angle rotations.
The initial learning rate is set to $5^{-5}$ and reduced by a factor of 10, when the 3D pose accuracy does not improve after every 5 epochs.
We train the network for 30 epochs with one NVIDIA RTX 2080Ti GPU.
PyTorch~\cite{paszke2017automatic} is used for code implementation.

\section{Experiment}

\subsection{Evaluation metrics and datasets.}

\noindent\textbf{Evaluation metrics.}
We report the per-frame and temporal evaluation metrics.
For the per-frame evaluation, we use mean per joint position error (MPJPE), Procrustes-aligned MPJPE (PA-MPJPE), and mean per vertex position error (MPVPE).
The position errors are measured in millimeter ($mm$) between the estimated and groundtruth 3D coordinates after aligning the root joint.
Particularly, we use PA-MPJPE as the main metric for per-frame accuracy, since it excludes the effect of outputs' scale ambiguity on errors.
For the temporal evaluation, we use the acceleration error proposed in HMMR~\cite{kanazawa2019learning}.
The acceleration error computes an average of the difference between the predicted and groundtruth acceleration of each joint in ($mm/s^2$).
% It indicates the temporal consistency and smoothness of produced 3D human motions.

\noindent\textbf{Datasets.}
We use 3DPW~\cite{von20183dpw}, Human3.6M~\cite{ionescu2014human3},
MPI-INF-3DHP~\cite{mehta2017monocular}, InstaVariety~\cite{kanazawa2019learning}, Penn Action~\cite{zhang2013actemes}, and PoseTrack~\cite{andriluka2018posetrack} for training, following VIBE~\cite{kocabas2020vibe}.
3DPW is the only in-the-wild dataset that contains accurate groundtruth SMPL parameters.
3DPW, Human3.6M, MPI-INF-3DHP are also used for evaluation.
More details are in the supplementary material.

\begin{figure}
\begin{center}\centering
\setlength{\tabcolsep}{0.04cm}
\setlength{\itemwidth}{2.55cm}
\hspace*{-\tabcolsep}
\begin{tabular}{ccc}

\animategraphics[width=\itemwidth, poster=32, loop, autoplay, final, nomouse, method=widget]{25}{fig/ablation/input/}{000010}{000055}
&
\animategraphics[width=\itemwidth, poster=32, loop, autoplay, final, nomouse, method=widget]{25}{fig/ablation/ours/}{000010}{000055}
&
\animategraphics[width=\itemwidth, poster=32, loop, autoplay, final, nomouse, method=widget]{25}{fig/ablation/baseline/}{000010}{000055}
\\

input
&
\textbf{-res +PF (Ours)}
&
+res -PF \\

% poster 후보: =15, =32
\end{tabular}

%\vspace{-1em}
\captionof{figure}
{
Qualitative comparison between our TCMR (middle) and the baseline (right).
TCMR learns more useful temporal features, and provides a more accurate 3D pose and temporally consistent 3D motion.
res denotes the residual connection and PF is the abbreviation for PoseForecast. 
% \emph{To watch the video, please refer to our arxiv paper.}
\emph{This is a video figure that is best viewed by Adobe Reader.}
}
\label{fig:ours_vs_base}
\end{center}
\vspace{-2.2em}
\end{figure}

\subsection{Ablation study}
In this study, we show how each component of our temporal architecture reduces the dependency of the model on a current static feature, and make it focus on temporal features from the past and future.
We take the same baseline used in VIBE~\cite{kocabas2020vibe}.
% to demonstrate the effectiveness of the proposed architecture on producing temporally consistent and smooth 3D human motion.
The baseline has a single bi-directional GRU that encodes temporal features from all input frames and a residual connection between the static and temporal features as VIBE.
It also predicts each 3D pose and shape for all input frames in a single feed-forward, but does not use the motion discriminator.
We use 3DPW~\cite{von20183dpw}, MPI-INF-3DHP~\cite{mehta2017monocular}, InstaVariety~\cite{kanazawa2019learning}, and Penn Action~\cite{zhang2013actemes} for training, and 3DPW for evaluation. % following~\cite{luo20203d}.

\noindent\textbf{Effectiveness of residual connection removal.}
To analyze the effect of the residual connection between the static and temporal features, we compare the models with and without it. 
As shown in Table~\ref{table:ablation}, removing the residual connection decreases the acceleration error significantly, which indicates a considerable improvement in temporal consistency and smoothness of 3D human motion.
This finding verifies that the identity mapping of the current static feature inside the residual connection hinders a model from learning meaningful temporal features.
Moreover, the increased temporal consistency of 3D motion improves the per-frame 3D pose accuracy.
Figure~\ref{fig:ours_vs_base} illustrates how the enhanced temporal consistency contributes to better per-frame 3D pose estimation.
The sudden change of poses, caused by the inaccurate 3D pose estimation on specific frames, is disappeared.
The above comparisons clearly validate the effectiveness of removing the residual connection in terms of both per-frame and temporal metrics.

\begin{table}
\small
\centering
\setlength\tabcolsep{1.0pt}
\def\arraystretch{1.1}
\caption{Comparison between different temporal architectures. All networks estimate only on the middle frame of the input sequence.}
\begin{tabular}{C{2.8cm}C{2.0cm}|C{1.8cm}C{1.2cm}}
\specialrule{.1em}{.05em}{.05em}
remove residual & PoseForecast & PA-MPJPE$\downarrow$ & Accel$\downarrow$ \\ \hline
\xmark & \xmark & 55.6 & 29.2 \\ 
\cellcolor{Gray}\xmark & \cellcolor{Gray}\cmark & \cellcolor{Gray}55.0 & \cellcolor{Gray}24.9 \\
\cmark & \xmark & 54.2 & 8.7 \\ 
\cellcolor{Gray}\hspace{1.27cm}\cmark\hspace{0.3cm} \textbf{(Ours)} & \cellcolor{Gray}\cmark & \cellcolor{Gray}\textbf{53.9} & \cellcolor{Gray}\textbf{7.7} \\
% \hspace{0.3cm}
\specialrule{.1em}{.05em}{.05em}
\end{tabular}
% \vspace*{-3mm}
% \vspace*{-5mm}
\label{table:ablation}
\end{table}

\begin{table}
\small
\centering
\setlength\tabcolsep{1.0pt}
\def\arraystretch{1.1}
\caption{Comparison between PoseForecast that takes a current frame and that does not take a current frame.}
\begin{tabular}{C{3.8cm}|C{2.0cm}C{1.2cm}}
\specialrule{.1em}{.05em}{.05em}
PoseForecast input & PA-MPJPE$\downarrow$ & Accel$\downarrow$ \\ \hline
w. current frame & \textbf{53.8} & 10.3 \\
\cellcolor{Gray}\textbf{wo. current frame (Ours)} & \cellcolor{Gray}53.9 & \cellcolor{Gray}\textbf{7.7} \\

\specialrule{.1em}{.05em}{.05em}
\end{tabular}
% \vspace*{-3mm}
\vspace*{-1.5em}
% \vspace*{-5mm}
\label{table:PoseForecast_input}
\end{table}

\begin{table}
\small
\centering
\setlength\tabcolsep{1.0pt}
\def\arraystretch{1.1}
\caption{Comparison between different supervision on estimated SMPL parameters from the PoseForecast.}
\begin{tabular}{C{4.8cm}|C{2.0cm}C{1.2cm}}
\specialrule{.1em}{.05em}{.05em}
PoseForecast supervision target & PA-MPJPE$\downarrow$ & Accel$\downarrow$ \\ \hline
none & 55.1 & 8.3 \\
\cellcolor{Gray}GT of past and future frames & \cellcolor{Gray}54.1 & \cellcolor{Gray}8.5\\
\textbf{GT of current frame (Ours)} & \textbf{53.9} & \textbf{7.7}\\

\specialrule{.1em}{.05em}{.05em}
\end{tabular}
% \vspace*{-3mm}
\vspace*{-1.5em}
% \vspace*{-5mm}
\label{table:PoseForecast_supervision}
\end{table}

\noindent\textbf{Effectiveness of PoseForecast}
We compare the models with and without PoseForecast to verify the effectiveness of forecasting current temporal features only from the past and future frames.
On the basis of the results in Table~\ref{table:ablation}, PoseForecast consistently improves per-frame and temporal metrics regardless of the residual connection. % of the existence
Particularly, the acceleration error consistently decreases by over 11\% .
Thus, the temporal encoding that takes all frames with the current frame may be suboptimal, and forecasting the current temporal features from the past and future frames is beneficial for temporally consistent 3D human motion.

\begin{table*}[h]
	\centering
	\caption{
	Evaluation of state-of-the-art methods on on 3DPW~\cite{von20183dpw}, MPI-INF-3DHP~\cite{mehta2017monocular}, and Human3.6M~\cite{ionescu2014human3}. All methods except HMMR~\cite{kanazawa2019learning} do not use Human3.6M SMPL parameters from Mosh~\cite{loper2014mosh}, but use 3DPW train set for training following MEVA~\cite{luo20203d}. The number of input frames are following the protocols of the papers.
	}
	\vspace*{-0.8em}
	\resizebox{\textwidth}{!}{%
		\begin{tabular}{l|cccc|ccc|ccc|c}
		
% 			\toprule
            \specialrule{.1em}{.05em}{.05em}
            
			& \multicolumn{4}{c}{3DPW} & \multicolumn{3}{c}{MPI-INF-3DHP} & \multicolumn{3}{c}{Human3.6M} & \multicolumn{1}{|c}{number of} \\
			\cmidrule(lr){2-5} \cmidrule(lr){6-8} \cmidrule(lr){9-11}
			method & PA-MPJPE $\downarrow$ & MPJPE $\downarrow$ & MPVPE $\downarrow$ & Accel $\downarrow$ & PA-MPJPE $\downarrow$ & MPJPE $\downarrow$ & Accel $\downarrow$ & PA-MPJPE $\downarrow$ & MPJPE $\downarrow$ & Accel $\downarrow$ & input frames \\
			
			\midrule
			
		    HMMR~\cite{kanazawa2019learning} & 72.6 & 116.5 & 139.3 & 15.2 & - & - & - & 56.9 & - & - & 20 \\
			\cellcolor{Gray}VIBE~\cite{kocabas2020vibe} & \cellcolor{Gray}57.6 & \cellcolor{Gray}91.9 & \cellcolor{Gray}- & \cellcolor{Gray}25.4 & \cellcolor{Gray}68.9 & \cellcolor{Gray}103.9 & \cellcolor{Gray}27.3 & \cellcolor{Gray}53.3 & \cellcolor{Gray}78.0 & \cellcolor{Gray}27.3 & \cellcolor{Gray}\textbf{16} \\
			MEVA~\cite{luo20203d} & 54.7 & 86.9 & - & 11.6 & 65.4 & \textbf{96.4} & 11.1 & 53.2 & 76.0 & 15.3 & 90 \\
			\cellcolor{Gray}\textbf{TCMR (Ours)} & \cellcolor{Gray}\textbf{52.7} & \cellcolor{Gray}\textbf{86.5} & \cellcolor{Gray}\textbf{103.2}& \cellcolor{Gray}\textbf{6.8} & \cellcolor{Gray}\textbf{63.5} & \cellcolor{Gray}97.6 & \cellcolor{Gray}\textbf{8.5} & \cellcolor{Gray}\textbf{52.0} & \cellcolor{Gray}\textbf{73.6} & \cellcolor{Gray}\textbf{3.9} & \cellcolor{Gray}\textbf{16} \\

            \specialrule{.1em}{.05em}{.05em}
% 			\bottomrule
			
		\end{tabular}%
	}
	\label{tab:sota_video}
\end{table*}{}

\begin{figure*}
\begin{center}
\includegraphics[width=1.0\linewidth]{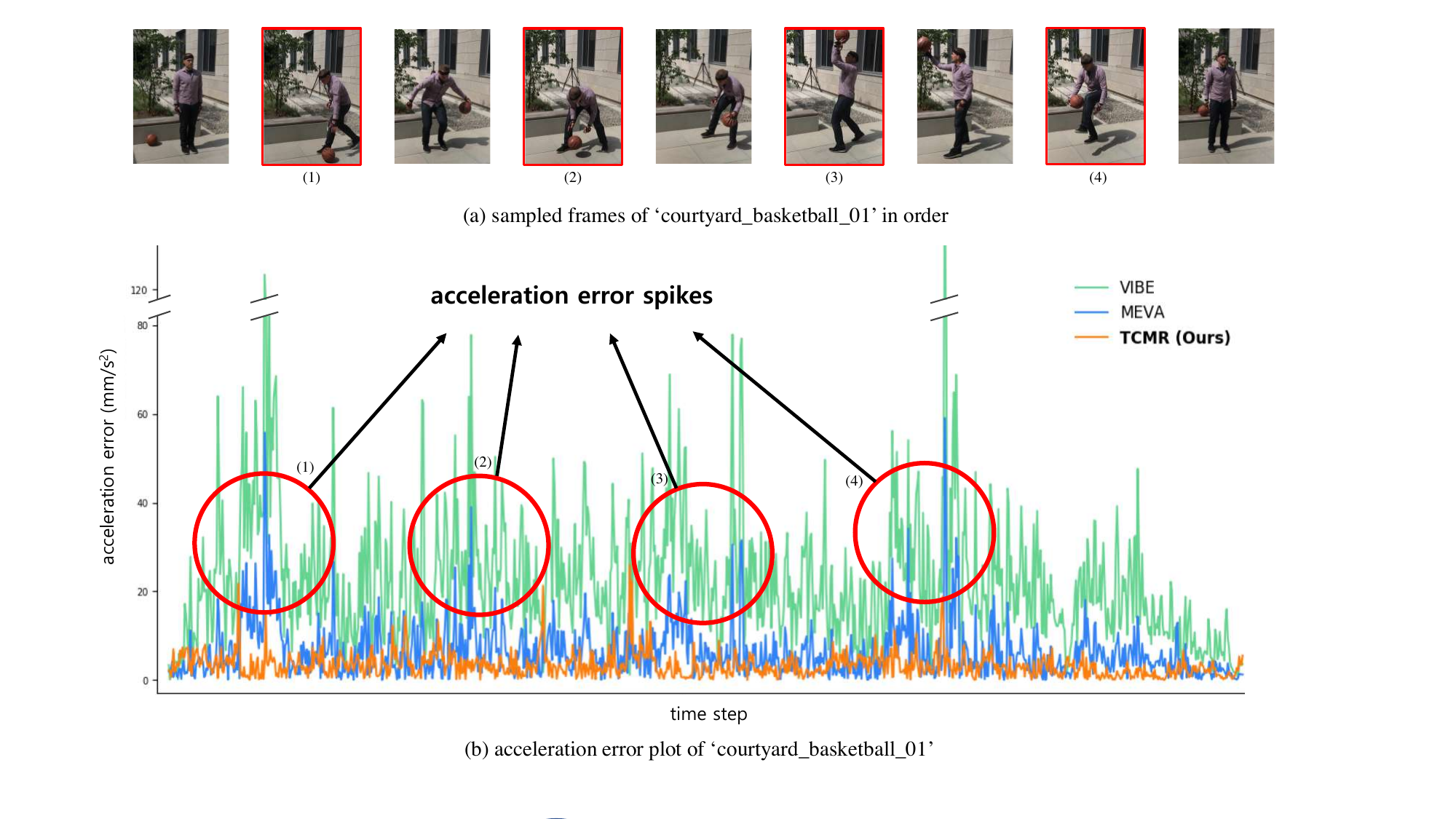}
\end{center}
\vspace*{-1em}
\caption{
Comparison between the acceleration errors of the proposed TCMR, MEVA~\cite{luo20203d}, and VIBE~\cite{kocabas2020vibe}. Our TCMR shows clearly lower acceleration errors along the time step than previous methods, which indicates temporally consistent 3D motion output. The previous methods reveal extreme acceleration error spikes compared to our TCMR.
}
\vspace*{-1.0em}
\label{fig:acceleration_plot}
\end{figure*}

To further validate the forecasting, we compare our PoseForecast with its variations.
First, we show the effectiveness of taking past and future frames without a current frame in Table~\ref{table:PoseForecast_input}.
As the table shows, additionally taking current frames increases the acceleration error by 33\%.
Thus, maintaining the temporal features free from the current static feature is important for temporally consistent and smooth 3D human motion.
% compare it with a temporal module that encodes additional current temporal features with a current static feature in Table~\ref{table:temporal_encoder}.
% It consists of two GRUs that cover $\mathbf{f}_1, \dots, \mathbf{f}_{\floor{T/2}}$ and $\mathbf{f}_{T}, \dots, \mathbf{f}_{\floor{T/2}}$ respectively.
% When the current static feature is also fed to the encoder, it focuses on the static feature and fails to fully exploit the temporal information.
% On the contrary, PoseForecast encodes the current temporal features free from the static feature, and thus drives the model to output more temporally consistent and smooth 3D human motion.
Second, we validate the effectiveness of supervising the predicted SMPL parameters from PoseForecast (\textit{i.e.}, $\Theta_\text{past}$ and $\Theta_\text{future}$) with groundtruth of the current frame in Table~\ref{table:PoseForecast_supervision}.
As shown in the table, supervising the predicted parameters with the current groundtruth provides better per-frame 3D pose accuracy and temporal consistency than the other supervisions.
When we supervise the predicted parameters with the groundtruth of $\floor{T/2}-1$th and $\floor{T/2}+1$th frames (the second row), the acceleration error increases by 10\%.
The performance degrades, because the temporal features of PoseForecast are encoded from the input including the static features of the target frames (\textit{i.e.}, $\floor{T/2}-1$th and $\floor{T/2}+1$th frames).
As verified in Table~\ref{table:PoseForecast_input}, including the target static feature hinders PoseForecast from learning useful temporal information for temporally consistent and smooth 3D human motion.
The encoded temporal feature is likely to be dominated by the target static feature and marginally leverage temporal information from other frames.
% We analyze that the performance degrades due a strong dependency of the PoseForecast on static features of the target frames (\textit{i.e.}, $\floor{T/2}-1$th and $\floor{T/2}+1$th frames).
% The strong dependency occurs because PoseForecast encodes the temporal features from the input including the static features of target frames (\textit{i.e.}, $\floor{T/2}-1$th and $\floor{T/2}+1$th frames).
% Including the target static features as input adversely affects to encode useful temporal feature for temporally consistent 3D motion, as verified in Table~\ref{table:PoseForecast_input}.
% Thus, PoseForecast tends to learn suboptimal temporal information due to the adverse effect of the dominance verified in Table~\ref{table:PoseForecast_input}.
%
% We think the performance degradation is from a strong dependency of the PoseForecast on static features of the target frames (\textit{i.e.}, $\floor{T/2}-1$th and $\floor{T/2}+1$th frames).
% As the target frames are included in the input temporal space, the supervision can make PoseForecast strongly dependent on the static features of the target frames, thus it fails to effectively encode temporal information.
% Table~\ref{table:PoseForecast_input} consistently shows that the system performs worse when the target frames are included in the input temporal space.
%
%This suggests that the temporal encoding is adversely affected by the potential dominance of the static feature of the $\floor{T/2}-1$th and $\floor{T/2}+1$th frames on the temporal features of PoseForecast.
When no supervision is observed (the first row), both 3D pose accuracy and temporal consistency decrease compared with ours.
Hence, designing our PoseFrecast to forecast the current SMPL parameters by supervising it with the current target (the third row) facilitates the network to learn more useful temporal features.

In summary, the above comparisons show that forecasting current temporal features from past and future frames is effective for temporally consistent 3D human motion by reducing the strong dependency on a current static feature.

\begin{table}
\small
\centering
\setlength\tabcolsep{1.0pt}
\def\arraystretch{1.1}
\caption{Comparison between ours and previous methods applied with average filtering on 3DPW~\cite{von20183dpw}.}
\vspace*{-0.5em}
% \resizebox{0.4\textwidth}{!}{%
    \begin{tabular}{L{2.5cm}|C{2.0cm}C{1.5cm}C{1.3cm}}
    \specialrule{.1em}{.05em}{.05em}
    method & PA-MPJPE$\downarrow$ & MPJPE$\downarrow$ & Accel$\downarrow$ \\ \hline
    VIBE~\cite{kocabas2020vibe} & 57.6 & 91.9 & 25.4  \\
    \cellcolor{Gray} + Avg. filter  & \cellcolor{Gray}57.8 & \cellcolor{Gray}91.6 & \cellcolor{Gray}13.5 \\ \hline
    MEVA~\cite{luo20203d} & 54.7 & 86.9 & 11.6  \\
    \cellcolor{Gray} + Avg. filter  & \cellcolor{Gray}55.5 & \cellcolor{Gray}87.7 & \cellcolor{Gray}8.2 \\ \hline
    \textbf{TCMR (Ours)} & \textbf{52.7} & \textbf{86.5} & 6.8 \\
    \cellcolor{Gray} + Avg. filter  & \cellcolor{Gray}55.0 & \cellcolor{Gray}88.7 & \cellcolor{Gray}\textbf{6.2} \\
    
    \specialrule{.1em}{.05em}{.05em}
    \end{tabular}
% }
\vspace*{-1.3em}
\label{table:average_filtering}
\end{table}

\begin{table*}[!t]
	\centering
	\caption{
	Evaluation of state-of-the-art methods on 3DPW~\cite{von20183dpw}, MPI-INF-3DHP~\cite{mehta2017monocular}, and Human3.6M~\cite{ionescu2014human3}. All methods do not use 3DPW~\cite{von20183dpw} on training. `single image' or `video' denotes whether the input of a method is a single image or a video.
	}
	\vspace*{-0.4em}
	\resizebox{\textwidth}{!}{%
		\begin{tabular}{ll|cccc|ccc|ccc}
		    
			\specialrule{.1em}{.05em}{.05em}

			&  & \multicolumn{4}{c}{3DPW} & \multicolumn{3}{c}{MPI-INF-3DHP} & \multicolumn{3}{c}{Human3.6M} \\
			\cmidrule(lr){3-6} \cmidrule(lr){7-9} \cmidrule(lr){10-12}
			method & & PA-MPJPE $\downarrow$ & MPJPE $\downarrow$ & MPVPE $\downarrow$ & Accel $\downarrow$ & PA-MPJPE $\downarrow$ & MPJPE $\downarrow$ & Accel $\downarrow$ & PA-MPJPE $\downarrow$ & MPJPE $\downarrow$ & Accel $\downarrow$ \\
			
			\midrule

			\parbox[t]{2mm}{\multirow{6}{*}{\rotatebox[origin=c]{90}{\hspace{0.4cm}single image}}}  & HMR~\cite{kanazawa2018end} & 76.7 & 130.0 & - & 37.4 & 89.8 & 124.2 & - & 56.8 & 88.0 & - \\
			& \cellcolor{Gray}GraphCMR~\cite{kolotouros2019convolutional} & \cellcolor{Gray}70.2 & \cellcolor{Gray}- & \cellcolor{Gray}- & \cellcolor{Gray}- & \cellcolor{Gray}- & \cellcolor{Gray}-& \cellcolor{Gray}- & \cellcolor{Gray}50.1 & \cellcolor{Gray}- & \cellcolor{Gray}- \\
			& SPIN~\cite{kolotouros2019learning} &  59.2 & 96.9 & 116.4 & 29.8 & 67.5 & 105.2 & - & \textbf{41.1} & - & 18.3 \\
			& \cellcolor{Gray}I2L-MeshNet~\cite{moon2020i2l} & \cellcolor{Gray}57.7 & \cellcolor{Gray}93.2 & \cellcolor{Gray}110.1 & \cellcolor{Gray}30.9 & \cellcolor{Gray}- & \cellcolor{Gray}- & \cellcolor{Gray}- & \cellcolor{Gray}\textbf{41.1} & \cellcolor{Gray}55.7 & \cellcolor{Gray}13.4 \\
			& Pose2Mesh~\cite{choi2020p2m} & 58.3 & \textbf{88.9} & \textbf{106.3} & 22.6 & - & - & - & 46.3 & 64.9 & 23.9 \\
			& \cellcolor{Gray}HKMR~\cite{georgakis2020hierarchical} & \cellcolor{Gray}- & \cellcolor{Gray}- & \cellcolor{Gray}- & \cellcolor{Gray}- & \cellcolor{Gray}- & \cellcolor{Gray}- & \cellcolor{Gray}- & \cellcolor{Gray}- & \cellcolor{Gray}59.6 & \cellcolor{Gray}- \\

			\midrule
			
		    \parbox[t]{2mm}{\multirow{5}{*}{\rotatebox[origin=c]{90}{video}}} 
		    & HMMR~\cite{kanazawa2019learning} & 72.6 & 116.5 & 139.3 & 15.2 & - & - & - & 56.9 & - & - \\
			&  \cellcolor{Gray}Doersch~\etal~\cite{doersch2019sim2real} & \cellcolor{Gray}74.7 & \cellcolor{Gray}- &  \cellcolor{Gray}- & \cellcolor{Gray}- & \cellcolor{Gray}- & \cellcolor{Gray}- &  \cellcolor{Gray}- &  \cellcolor{Gray}- & \cellcolor{Gray}- & \cellcolor{Gray}- \\
			& Sun~\etal~\cite{sun2019human} & 69.5 & - & - & - & - & - & - & 42.4 & \textbf{59.1} & - \\
			& \cellcolor{Gray}VIBE~\cite{kocabas2020vibe} & \cellcolor{Gray}56.5 & \cellcolor{Gray}93.5 & \cellcolor{Gray}113.4 & \cellcolor{Gray}27.1 & \cellcolor{Gray}63.4 & \cellcolor{Gray}97.7 & \cellcolor{Gray}29.0 & \cellcolor{Gray}41.5 & \cellcolor{Gray}65.9 & \cellcolor{Gray}18.3 \\
			& \textbf{TCMR (Ours)} & \textbf{55.8} & 95.0 & 111.3 & \textbf{6.7} & \textbf{62.8} & \textbf{96.5} & \textbf{9.5} & \textbf{41.1} & 62.3 & \textbf{5.3} \\ 
			
            \specialrule{.1em}{.05em}{.05em}
            
		\end{tabular}%
	}
	\label{tab:sota}
\vspace*{-1.0em}
\end{table*}{}

\subsection{Comparison with state-of-the-art methods}
\noindent\textbf{Comparison with video-based methods.}
We compare our TCMR with previous state-of-the-art video-based methods ~\cite{kanazawa2019learning,kocabas2020vibe,luo20203d} that report the acceleration error in Table~\ref{tab:sota_video}.
On the basis of the study of Luo~\etal~\cite{luo20203d}, all methods, except HMMR~\cite{kanazawa2019learning} are trained on the train set including 3DPW~\cite{von20183dpw}, but do not leverage Human3.6M~\cite{ionescu2014human3} SMPL parameters obtained from Mosh~\cite{loper2014mosh} for supervision.
The numbers of VIBE~\cite{kocabas2020vibe} are from MEVA~\cite{luo20203d}, but we validated them independently.
As shown in the table, our proposed system outperforms the previous video-based methods on all benchmarks both in per-frame 3D pose accuracy and temporal consistency.
These results prove that our system effectively leverages temporal information of the past and future by resolving the system's strong dependency on a current static feature.
% Moreover, our system does not suffer from the trade-off between accuracy and smoothness.
% As discussed in VIBE~\cite{kocabas2020vibe}, HMMR~\cite{kanazawa2019learning} produces smoother predictions than VIBE, but tends to sacrifice the per-frame accuracy by over-smoothing 3D motion.
% On the other hand, our method not only provides more temporally consistent 3D motion, but also more accurate per-frame 3D poses.
Although MEVA~\cite{luo20203d} also improves the per-frame and temporal metrics, the model consumes nearly 6 times more input frames during training and testing, and provides worse results than ours.
In addition, MEVA requires at least 90 input frames, which means that it can not be trained and tested on short videos.
Figure~\ref{fig:acceleration_plot} describes the clear advantage of our TCMR on the temporal consistency among video-based methods.
The previous methods expose numerous spikes, which represent unstable and unsmooth 3D motion estimation. 
Our TCMR provides relatively low acceleration errors along the time step, which indicates temporally consistent 3D motion output.
The figure's acceleration errors are measured on a sequence of the 3DPW validation set that has a diverse motion.

To further confirm the effectiveness of the proposed system on temporal consistency, we compare our TCMR with VIBE~\cite{kocabas2020vibe} and MEVA~\cite{luo20203d} with an average filter applied as post-processing in Table~\ref{table:average_filtering}.
Average filtering is performed by spherical linear interpolation in the quaternions of estimated SMPL~\cite{loper2015smpl} pose parameters following MEVA.
The numbers of other methods are from MEVA.
As shown in the table, our system outperforms other methods even when they are applied with the average filtering.
%In addition, we can observe the trade-off between per-frame accuracy and temporal consistency from the result of MEVA, when it is applied with the average filtering .
%This implies that filtering could smooth out the details of 3D human motion due to its low-pass filtering mechanism; thus it can worse per-frame 3D pose accuracy.
% We found that 
Moreover, the results imply that the average filtering can decrease the per-frame 3D pose accuracy by smoothing out the details of 3D human motion.
However, each component of our TCMR decreases the acceleration error while improving the per-frame 3D pose accuracy, as shown in Table~\ref{table:ablation}.
% indicates that since filtering is fundamentally based on a low-pass filter, it could smooth out the details of 3D human motions.

In summary, our newly designed system significantly outperforms the previous state-of-the-art methods in temporal consistency and smoothness of 3D human motion without any post-processing while also increasing the per-frame 3D pose accuracy.
Note that the comparison in Table~\ref{tab:sota_video} and~\ref{table:average_filtering} is the fairest comparison between the video-based methods, since all methods, except HMMR~\cite{kanazawa2019learning}, used the same training datasets.

\noindent\textbf{Comparison with single image-based and video-based methods.}
We compare our system with previous 3D pose and shape estimation methods, including single image-based methods in Table~\ref{tab:sota}.
None of the methods are trained on 3DPW~\cite{von20183dpw}.
For evaluation on Human3.6M~\cite{ionescu2014human3}, we use the frontal view images following~\cite{kanazawa2019learning,kolotouros2019learning}, whereas all views are tested in Table~\ref{tab:sota_video} and~\ref{table:average_filtering}. 
In addition, to confirm the acceleration error of VIBE~\cite{kocabas2020vibe} on MPI-INF-3DHP~\cite{mehta2017monocular} and Human3.6M, we re-evaluate the model using the pretrained weights provided in the official code repository.

As shown in the table, our method outperforms all the previous methods on 3DPW, a challenging in-the-wild benchmark, and MPI-INF-3DHP in per-frame 3D pose accuracy (PA-MPJPE) and temporal consistency.
Especially the temporal consistency is largely improved compared with single image-based methods.
While VIBE decreases the acceleration error of SPIN~\cite{kolotouros2019learning} by 9\% and is defeated by Pose2Mesh~\cite{choi2020p2m} in the temporal consistency, our system provides over 3 times better performance than both SPIN and Pose2Mesh in 3DPW.
Moreover, VIBE gives a higher acceleration error than I2L-MeshNet~\cite{moon2020i2l} but our TCMR outperforms it by a wide margin in Human3.6M.

We provide qualitative comparison with VIBE~\cite{kocabas2020vibe} and MEVA~\cite{luo20203d} on 3DPW, qualitative results of TCMR on Internet videos, and failure cases in this link \footnote{\url{https://www.youtube.com/watch?v=WB3nTnSQDII}}.

% \vspace*{-0.4em}
\section{Conclusion}
% \vspace*{-0.2em}
We present TCMR, a novel and powerful system that estimates a 3D human mesh from a RGB video.
Previous video-based methods suffer from the temporal inconsistency issue because of the strong dependency on the static feature of the current frame.
We resolve the issue by removing the residual connection between the static and temporal features, and employing PoseForecast that forecasts the current temporal feature from the past and future frames.
In comparison with the previous video-based methods, the proposed TCMR provides highly temporally consistent 3D motion and a more accurate 3D pose per frame.

\vspace*{+2mm}

\noindent\textbf{Acknowledgements.}
% \section*{Acknowledgements.}
This work was supported by IITP grant funded by the Ministry of Science and ICT of Korea (No. 2017-0-01780) and AIRS Company in Hyundai Motor Company \& Kia Motors Corporation through HKMC-SNU AI Consortium Fund.

\clearpage

\twocolumn[{
\begin{center}
\begin{Large}
\textbf{Supplementary Material \textit{for} \\ \vspace{2mm}
Beyond Static Features for Temporally Consistent\\3D Human Pose and Shape from a Video}

\end{Large}
\end{center}
\vspace*{+2em}
}]

\section{More qualitative results}
% \vspace*{-1.5mm}
We provide more qualitative results in the online video \footnote{\url{https://www.youtube.com/watch?v=WB3nTnSQDII}}, which consists of three parts.
The first part shows the qualitative results of our TCMR on in-the-wild videos that have fast and diverse motions from 3DPW~\cite{von20183dpw}.
We also provide the outputs rendered from the opposite view.
The second part compares the proposed TCMR with VIBE~\cite{kocabas2020vibe} and MEVA~\cite{luo20203d}.
The results are rendered on a plain background with a fixed camera to clearly compare the temporal consistency and smoothness of 3D human motion following MEVA~\cite{luo20203d}.
The fixed camera has the fixed weak-perspective camera parameters $s$ and $t$, which are set to one and zero, respectively.
The last part provides the results of TCMR on Internet videos.
The bounding boxes of people in the videos are tracked by a multi-person tracker that uses YOLOv3~\cite{redmon2018yolov3}.
With the cropped images from the bounding boxes, our TCMR processes 41 frames per second (fps) for the video \footnote{\url{https://www.youtube.com/watch?v=Opry3F6aB1I}} with 5 people. 
A single NVIDIA RTX 2080Ti GPU is used for the test.

% \vspace*{-1.5mm}
\section{Human evaluation.}
% \vspace*{-1.5mm}
We surveyed 50 people to pick the most realistic motion from
TCMR, MEVA, and VIBE outputs on 20 sequences of 3DPW~\cite{von20183dpw} validation and test sets.
TCMR, MEVA, and VIBE got 69\%, 26\%, and 5\% votes, respectively.
The result is coherent with the acceleration error results of the three methods in the main manuscript.

% \vspace*{-1.5mm}
\section{Attention values in feature integration.}
% \vspace*{-1.5mm}
During the temporal feature integration, the past and future temporal features are weighted more than the current temporal feature, and the variation range of each attention value is $\pm$20\%.
The past and future temporal features' attention values tend to become larger when the current pose is difficult or the motion is fast. 
The attached videos\footnote{\url{https://youtu.be/dFQ6hkfkwz0}}\footnote{\url{https://youtu.be/otdL5WVjwPg}} plot the attention values of the past, future, and current temporal features on two sequences of 3DPW~\cite{von20183dpw}.
The values are written at the top-right of frames, and the sum is always $1$. 
As the video shows, the attention value of the current temporal feature does not drop below $0.4$ when a subject is walking in slow motion, whereas the value overall stays below $0.4$ when a subject is playing basketball with fast movement and complex poses.

% \vspace*{-1.5mm}
\section{Datasets}
% \vspace*{-1.5mm}
\noindent\textbf{3DPW.}
3DPW~\cite{von20183dpw} is captured from in-the-wild and contains 3D human pose and shape annotations.
It consists of 60 videos and 51K video frames in total, which are captured with a phone at 30 fps.
IMU sensors are leveraged to acquire the groundtruth 3D human pose and shape.
We follow the official split protocol to train and test our model, where train, validation, test sets consist of 24, 12, 24 videos, respectively.
Also, we report MPVPE on 3DPW because it only has groundtruth 3D shape among the datasets we used.
We use 14 joints defined by Human3.6M~\cite{ionescu2014human3} for evaluating PA-MPJPE and MPJPE following the previous works~\cite{kanazawa2018end,kanazawa2019learning,kolotouros2019learning,kocabas2020vibe}.

\noindent\textbf{Human3.6M.}
Human3.6M~\cite{ionescu2014human3} is a large-scale indoor 3D human pose benchmark, which consists of 15 action categories and 3.6M video frames. 
% The groundtruth 3D human poses are obtained using a motion capture system, but there are no groundtruth SMPL~\cite{loper2015smpl} parameters.
% Accordingly, most of the previous 3D human pose and shape estimation works~\cite{kanazawa2018end,kolotouros2019convolutional,kolotouros2019learning,kocabas2020vibe} used pseudo-groundtruth obtained from Mosh~\cite{loper2014mosh}.
% However, because of the license issue, the pseudo-groundtruth from Mosh is not currently publicly accessible.
% Thus, we generate new pseudo-groundtruth SMPL parameters by fitting SMPL parameters to the groundtruth 3D human poses.
Following~\cite{kocabas2020vibe}, our TCMR is trained on 5 subjects (S1, S5, S6, S7, S8) and tested on 2 subjects (S9, S11).
We subsampled the dataset to 25 fps (originally 50 fps) for training and evaluation on the acceleration error.
14 joints defined by Human3.6M are used for computing PA-MPJPE and MPJPE.

\begin{table*}[!t]
\centering
\caption{Comparison between different models using ResNet with different initialization to extract static features. All models use the same SMPL parameter regressor pretrained by SPIN~\cite{kolotouros2019learning}.}
\vspace*{-3mm}
\begin{tabular}{C{8.3cm}C{2.1cm}C{2.1cm}|C{1.8cm}C{1.2cm}}
\specialrule{.1em}{.05em}{.05em}
ResNet initialization  & remove residual & PoseForecast & PA-MPJPE$\downarrow$ & Accel$\downarrow$ \\ \hline
ResNet with random initialization & \xmark & \xmark & 126.5 & 24.3 \\
\cellcolor{Gray}ResNet pretrained on ImageNet~\cite{russakovsky2015imagenet} & \cellcolor{Gray}\xmark & \cellcolor{Gray}\xmark & \cellcolor{Gray}103.7 & \cellcolor{Gray}65.5\\
ResNet from SPIN~\cite{kolotouros2019learning} & \xmark & \xmark & 55.6 & 29.2\\
\cellcolor{Gray}\textbf{\hspace{0.3cm}ResNet from SPIN~\cite{kolotouros2019learning}\hspace{0.1cm} (TCMR. Ours.)} & \cellcolor{Gray}\cmark & \cellcolor{Gray}\cmark & \cellcolor{Gray}\textbf{53.9} & \cellcolor{Gray}\textbf{7.7}\\

\specialrule{.1em}{.05em}{.05em}
\end{tabular}
\vspace*{-3mm}

\label{table:pretrain_ablation}
\end{table*}

\noindent\textbf{MPI-INF-3DHP.}
MPI-INF-3DHP~\cite{mehta2017monocular} is a 3D benchmark mostly captured from indoor environment.
The train set has 8 subjects, 16 videos per subject, and 1.3M video frames captured at 25 fps in total.
It exploits a marker-less motion capture system and provides 3D human pose annotations.

The test set contains 6 subjects performing 7 actions in both the indoor and outdoor environment.
The positional errors (\textit{i.e.}, PA-MPJPE and MPJPE) of TCMR are measured on the valid frames, which are composed of every 10th frame approximately, using 17 joints defined by MPI-INF-3DHP.
The acceleration error is computed using all frames.

\noindent\textbf{InstaVariety.}
InstaVariety is a 2D human dataset curated by HMMR~\cite{kanazawa2019learning}, whose videos are collected from Instagram using 84 motion-related hashtags.
There are 28K videos with an average length of 6 seconds, and OpenPose~\cite{cao2017realtime} is used to acquire pseudo-groundtruth 2D pose annotations.

\noindent\textbf{Penn Action.}
Penn Action~\cite{zhang2013actemes} contains 2.3K video sequences of 15 different sports actions.
It has a total of 77K video frames annotations for 2D human poses, bounding boxes, and action categories.

\noindent\textbf{PoseTrack.}
PoseTrack~\cite{andriluka2018posetrack} is a 2D benchmark for multi-person pose estimation and tracking in videos.
It contains 1.3K videos and 46K annotated frames in total.
The videos are captured at different fps, varying around 25 fps.
We use 792 videos from the official train set, which has 2D pose annotations for 30 frames in the middle of the video.

\section{Effect of pretrained ResNet}

Due to lack of video data, our TCMR and previous video-based methods~\cite{kanazawa2019learning,kocabas2020vibe,luo20203d} employ ResNet~\cite{he2016deep} pretrained by the single image-based 3D human pose and shape estimation methods~\cite{kanazawa2018end,kolotouros2019learning} to extract static features from input frames.
The pretrained ResNet is trained on large-scale in-the-wild 2D human pose datasets and provides reliable static features.
However, it is also one reason for the strong dependency of the system on the current static feature.
The current static feature extracted by the pretrained ResNet already contains a strong cue on the current 3D human pose and shape, leading the system to leverage temporal information marginally.

In this regard, an alternative to our TCMR, one could train models from scratch without using the ResNet pretrained by~\cite{kanazawa2018end,kolotouros2019learning} to extract static features to reduce the strong dependency.
Table~\ref{table:pretrain_ablation} compares our TCMR, the baseline (the third row), and the models that do no use the ResNet pretrained by SPIN~\cite{kolotouros2019learning}.
As the table shows, the models that do no use the ResNet pretrained by SPIN~\cite{kolotouros2019learning} reveal very high per-frame 3D pose errors.
This indicates that training the models with only video data in the current literature is not sufficient for accurate 3D human pose estimation.
The interesting part is that the model using ResNet with random initialization provides the highest 3D pose error but the lowest acceleration error among the models without our TCMR.
While the high pose error attributes to the lack of train data, the low acceleration error implies that the strong cue of the current static feature adversely affects the temporal consistency of 3D human motion.

In summary, with the insufficient video data in the current literature, the proposed TCMR significantly improves the temporal consistency of 3D human motion by reducing the strong dependency on the current static feature.
It also preserves the per-frame 3D pose accuracy by leveraging the ResNet pretrained on large-scale in-the-wild 2D human pose datasets to extract useful static features.

\section{Effect of input fps}

\begin{table}
\small
\centering
\setlength\tabcolsep{1.0pt}
\def\arraystretch{1.1}
\caption{Performance comparison between two networks taking different input fps on 3DPW~\cite{von20183dpw}. The numbers in the second row are from Table $4$ of the main manuscript.}
\begin{tabular}{C{2.2cm}|C{2.5cm}C{1.5cm}}
\specialrule{.1em}{.05em}{.05em}
input fps & PA-MPJPE$\downarrow$ & Accel$\downarrow$ \\ \hline
15 & 53.5 & 15.3 \\
\cellcolor{Gray}\textbf{30} & \cellcolor{Gray}\textbf{52.7} & \cellcolor{Gray}\textbf{7.1} \\

\specialrule{.1em}{.05em}{.05em}
\end{tabular}
\label{table:fps_ablation}
\vspace*{-3mm}
\end{table}

Table~\ref{table:fps_ablation} shows the effect of input fps.
The acceleration error doubles when input fps reduces by half, whereas the accuracy remains relatively the same.
The result indicates that TCMR can still fix invalid poses using relatively sparse temporal information.
The result also implies that temporally dense information is critical for temporal consistency of outputs, which is intuitive.

\section{Pose2Mesh with temporal smoothing}

We performed temporal smoothing on Pose2Mesh~\cite{choi2020p2m}, the state-of-the-art single image-based 3D human pose and shape estimation method.
Pose2Mesh wins the first in MPJPE, MPVPE, and acceleration error and the second in PA-MPJPE among single image-based methods according to Table 6 of the main manuscript.
Pose2Mesh with euro-filter achieves PA-MPJPE 58.6, MPJPE 89.6, acceleration error 12.9 on 3DPW. 
TCMR still outperforms the smoothed Pose2Mesh by nearly twice in temporal consistency without any post-processing.

\clearpage
{\small
\bibliographystyle{ieee_fullname}
\bibliography{main}
}

\end{document}